\documentclass[conference]{IEEEtran}
\usepackage{cite}
\usepackage{amsmath,amssymb,amsfonts}
\usepackage{algorithmic}
\usepackage{graphicx}
\usepackage{textcomp}
\usepackage{comment}
\usepackage{xcolor}
\usepackage{subcaption}
\usepackage{amsmath}
\usepackage{amssymb}
\usepackage{gensymb}
\usepackage{multirow}
\usepackage{array}
\usepackage{bm}
\usepackage[left=1.62cm,right=0.64in,top=0.75in]{geometry}

\usepackage{array}
\usepackage{lipsum}
\usepackage{booktabs}
\usepackage{balance}

\def\BibTeX{{\rm B\kern-.05em{\sc i\kern-.025em b}\kern-.08em
    T\kern-.1667em\lower.7ex\hbox{E}\kern-.125emX}}
\begin{document}

\title{Optimizing UAV Aerial Base Station Flights Using DRL-based Proximal Policy Optimization }

\author{
\IEEEauthorblockN{Mario Rico Ibáñez$^1$; Azim Akhtarshenas$^1$; David L\'opez-P\'erez$^1$; Giovanni Geraci$^2$}
\IEEEauthorblockA{\textit{$^1$Universitat Politècnica de València, Valencia, Spain}} 
\IEEEauthorblockA{\textit{$^2$Telefónica Research and Universitat Pompeu Fabra, Barcelona, Spain}} 
\textit{mriciba@upv.edu.es}}

\maketitle

\begin{abstract}

\footnotetext{This research is supported by the Generalitat Valenciana
through the CIDEGENT PlaGenT, Grant CIDEXG/2022/17, Project iTENTE, and the action CNS2023-144333, financed by
MCIN/AEI/10.13039/501100011033 and the European Union
“NextGenerationEU”/PRTR.}
Unmanned aerial vehicle (UAV)-based base stations offer a promising solution in emergencies where the rapid deployment of cutting-edge networks is crucial for maximizing life-saving potential. 
Optimizing the strategic positioning of these UAVs is essential for enhancing communication efficiency. 
This paper introduces an automated reinforcement learning approach that enables UAVs to dynamically interact with their environment and determine optimal configurations. 
By leveraging the radio signal sensing capabilities of communication networks, 
our method provides a more realistic perspective, 
utilizing state-of-the-art algorithm ---proximal policy optimization--- to learn and generalize positioning strategies across diverse user equipment (UE) movement patterns. 
We evaluate our approach across various UE mobility scenarios, including static, random, linear, circular, and mixed hotspot movements.
The numerical results demonstrate the algorithm's adaptability and effectiveness in maintaining comprehensive coverage across all movement patterns.
 
\textbf{UAV, Aerial Base Station, DRL, PPO, Sensing}
\end{abstract}


\section{Introduction} 

Effective communication during disasters is crucial for successful search and rescue operations. 
However, this can be particularly challenging when existing networks are compromised, 
leaving user equipment (UE) in urgent need of assistance. 
One innovative approach to address this issue is the use of unmanned aerial vehicles (UAVs),
which offers a practical solution for quickly establishing emergency communication networks~\cite{uav_emergency}. 
UAVs, acting as flying base stations (BSs), can play a critical role in scenarios where most ground BSs are overloaded, collapsed, or entirely absent. 
Such situations were evident during events like 9/11~\cite{11s_comm}, Typhoon Haiyan~\cite{aquino2013typhoon}, and in many urban, suburban, and rural search and rescue operations. 
In these critical circumstances, search and 
the rapid deployment of broadband and/or ultra-reliable low-latency communication (URLLC) networks is essential to support first responders as quickly as possible, 
maximizing the chances of saving lives.

\subsection{Related Works}

Extensive recent work on UAV-based emergency communication networks has primarily focused on UAV positioning optimization, 
which we discuss along with their drawbacks below.

The author in~\cite{cicek2019uav} surveyed UAV-BS location optimization,
introduced a generic mixed-integer non-linear programming (MINLP) framework,
proposed a taxonomy of solutions,
and outlined future research directions for 5G and beyond.
In \cite{8038869}, 
the authors developed a framework to optimize 3D UAV placement, mobility, device-to-UAV association, and uplink power control for IoT devices, 
aiming to minimize energy consumption while ensuring network connectivity.
Unfortunately, all these works lack real-time decision-making capabilities due to their static nature and their inability to model the complexity of real-world environments,
which are inherently non-convex, non-linear, and stochastic.

As a promising alternative, 
the integration of artificial intelligence (AI)---particularly reinforcement learning (RL)---into UAV networks enables autonomous, real-time optimization of 3D positioning, flight paths, and resource management.
For instance, the authors in \cite{8877247} addressed emergency scenarios by proposing a Q-learning-based solution to optimize the 3D positioning and transmit power of a network with multiple UAV-based small cells,
with the objective of maximizing network coverage while accounting for UE mobility.
Similarly, the study in \cite{ghanavi2018efficient} improved terrestrial networks using an aerial BS,
aiming to enhance quality of service (QoS) by mitigating performance degradation due to UE mobility.
This work employed Q-learning to determine the optimal aerial BS placement based on past experiences.
However, Q-learning suffers from the curse of dimensionality, 
making it impractical for large-scale problems wherein the state-action space is vast—as is the case in our scenario, 
making deep RL (DRL) a more suitable alternative where the Q-table is approximated by a deep neural network (DNN).

Assuming simple UE mobility scenarios, 
the authors in \cite{8432464} developed a DRL-based approach for UAV flight control,
aiming to maximize energy efficiency while balancing network coverage and UE data rates.
%
%
Ding~\textit{et al.} in \cite{ding20203d} modeled UAV energy consumption as a function of 3D movement,
and designed a DRL algorithm using deep deterministic policy gradient (DDPG) 
to optimize flight direction and speed for throughput maximization while satisfying energy constraints.
%
%
Focusing on convergence efficiency, 
the authors in \cite{parvaresh2023continuous} proposed a continuous actor-critic DRL (ACDRL) approach,
which reduced UAV-BS placement convergence time by 85\,\% compared to traditional DRL-based methods.
These studies, nonetheless, rely on discrete state-action spaces, often leading to suboptimal performance.
More critically, the above-mentioned DRL methods suffer from training instability due to high policy update variance and hyperparameter sensitivity,
limiting their effectiveness in dynamic environments.

To address this issue, 
trust region policy optimization (TRPO) was introduced,
enforcing a strict KL-divergence constraint to ensure stable policy updates~\cite{schulman2015trust}.
For instance, the authors in \cite{ho2021uav} applied TRPO to address instabilities in DDPG,
improving both training stability and learning efficiency.
Their approach optimized UAV control for energy efficiency and data rate maximization in dynamic wireless environments,
enhancing UAV-based wireless service provisioning.
However, TRPO’s KL-divergence hard constraint makes it computationally expensive.

Proximal policy optimization (PPO)  overcomes TRPO’s limitations by replacing its strict KL-divergence constraint with a clipped objective function, 
improving both efficiency and computational simplicity~\cite{schulman2017proximal}.
Saxena~\textit{et al.} in \cite{saxena2019optimal} introduced flow-level models (FLM) to evaluate UAV-BS network performance across multiple metrics,
and proposed a PPO-based approach to optimize traffic-aware UAV trajectories.
Their offline training aimed to maximize cumulative performance metrics,
while online simulations demonstrated a three-fold increase in average UE throughput and more balanced BS traffic loads compared to initial UAV placements.
Unfortunately, their method assumes perfect knowledge of UE positions,
which limits its applicability in real-world environments where UE mobility and uncertainty must be considered.
\cite{li2020trajectory} was among the first works to propose a PPO-based approach for UAV trajectory design, 
optimizing the sum rate for all UEs while considering their movement.
Their PPO-based solution outperformed Q-learning when UAVs followed specific paths
where UE velocity and distribution remained relatively stable.
To further adapt to dynamic UE distributions, 
they introduced a random-training algorithm (RT-PPO).
However, their approach restricts UAV locations to a set of predefined points,
which simplifies the problem but limits flexibility.

%
\subsection{Motivation and Contributions}

Although previous studies on UAV positioning have provided valuable insights,
many rely on unrealistic assumptions, including static UEs, known UE positions, predefined UAV paths, and discrete UAV movements.
These assumptions fail to capture the stochastic nature of UE mobility, the uncertainty of UE locations, and the flexibility of UAVs to navigate—especially in emergency scenarios.
As a result, existing approaches lack adaptability to dynamic environments, limiting their real-world applicability.

To address these challenges, we leverage PPO~\cite{schulman2017proximal}
to develop a more realistic and adaptive framework for UAV trajectory planning.
Specifically, we eliminate state-action space discretization and replace GPS-based positioning
with measurements over practical reference signals, 
allowing UAVs to position themselves based on real-time UE signal characteristics.

Our key contributions are:
\begin{itemize}
\item Enabling flexible UAV movement to accurately model continuous trajectory adjustments in real-world deployments.
\item Utilizing UE reference signals and angle of arrival (AOA) measurements, instead of GPS data, improving robustness in scenarios with unknown UE locations.
\item Evaluating performance across diverse UE mobility patterns to validate PPO’s effectiveness in dynamic emergency scenarios.
\end{itemize}

\section{System Model}
\label{sec:system_model}

\subsection{System Description}

In this paper, 
we consider an emergency cellular network comprised of multiple UAVs serving as mobile BSs,
which attempt to serve the UEs of a team of first responders. 
This cellular network operates in both downlink and uplink using time division multiple access (TDMA). 

Let $\mathcal{U} = \{u_{\mathit{1}}, \ldots,u_{\mathit{u}}, \ldots, u_{U}\}$, represent the set of first responders, 
each one with his/her own UE, 
and its cardinality (i.e., the number of elements in the set $\mathcal{U}$) be denoted by $U$.  
The geographical position of the first $u^{th}$ responder is determined by
$\rho^\mathrm{U}_u = \{x^\mathrm{U}_u, y^\mathrm{U}_u, z^\mathrm{U}_u\}$,
and the position matrix of all first responders is represented as 
$\bm{\rho}^\mathrm{U} = \{\rho^\mathrm{U}_\mathit{1}, \ldots, \rho^\mathrm{U}_\mathit{u}, \ldots,  \rho^\mathrm{U}_U\}$. 

Let $\mathcal{D} = \{d_1 , . . . , d_d , . . . , d_D\}$, denote the set of drones or UAVs, 
each one with its own BS, 
with its cardinality denoted by $D$. So, the location information of the $d^{th}$ UAV is represented by 
$\rho^\mathrm{D}_d = \{x^\mathrm{D}_d, y^\mathrm{D}_d, z^\mathrm{D}_d\}$,
and the position matrix of all UAVs is represented as 
$\bm{\rho}^\mathrm{D} = \{\rho^\mathrm{D}_1,...,\rho^\mathrm{D}_d,...,\rho^\mathrm{D}_D\}$.

\subsection{Channel Model}

We assume the network operates with a bandwidth $B$ at a frequency $f$. 
All radio links in the network experience slow and fast channel gains. 
We denote by $G_{u,d,k}$ the overall channel gain between the $u^{th}$ UE and the BS of the $d^{th}$ UAV in the $k^{th}$ frequency resource. 
This gain can be further decomposed into antenna gain ($G^{\rm a}$), path gain ($G^{\rm p}$), outdoor-to-indoor gain ($G^{\rm e}$), shadow gain ($G^{\rm s}$), and fast-fading gain ($G^{\rm{ff}}$),
i.e.
\begin{equation}
    G_{u,d,k} = G^{\rm a}_{u,d} \cdot G^{\rm p}_{u,d} \cdot G^{\rm e}_{u,d} \cdot G^{\rm s}_{u,d} \cdot \left|G^{\rm ff}_{u,d,k}\right|^2.
\end{equation}
In this study, 
we use the Urban Macro models specified by the third generation partnership project (3GPP) in TR25.814 to drive the calculation of each one of the presented channel components, 
with the following amendments: 
BS antennas are considered to be omnidirectional and the multi-path fading follows a Rician model. 
It is worth noting that the overall channel gain $G_{u,d,k}$ depends on both the position of the UAV $\rho^\mathrm{D}_d$ (our optimization variable) and the position of the first responder $\rho^\mathrm{U}_u$,
where this last one is unknown to us. 
Thus, the channel gain $G_{u,d,k}$  can be represented as 
$G_{u,d,k}(\rho^\mathrm{U}_u,\rho^\mathrm{D}_d)$.

The power received by the $u^{th}$ first responder from the BS of the $d^{th}$ UAV  within the $k^{th}$ frequency resource can be expressed as~\cite{meyers1946nonlinearity}:  
\begin{equation}
    P^{\rm{rx}}_{u,d,k}(\rho^{\mathrm{U}}_u, \rho^{\mathrm{D}}_d) = P^{\rm{tx}}_{d,k} \cdot G_{u,d,k}(\rho^{\mathrm{U}}_u, \rho^{\mathrm{D}}_d),
    \label{eq_intro:Received_power}
\end{equation}
where $P^{\rm{tx}}_d$ is the power transmitted by the BS of the $d^{th}$ UAV in the $k^{th}$ frequency resource.

The parameter that indicates the quality of the signal received by the $u^{th}$ first responder from the BS of the $d^{th}$ UAV in the $k^{th}$ frequency resource is the signal-to-interference-plus-noise ratio (SINR) $\gamma_{u,k}$, 
and can be calculated as~\cite{mozaffari2015drone, zhang2018downlink}:
\begin{eqnarray}
 	\gamma_{u,d,k}(\rho^{\mathrm{U}}_u, \bm{\rho}^\mathrm{D}) = \frac{P^{\mathrm rx}_{u,d,k}(\rho^{\mathrm{U}}_u, \rho^{\mathrm{D}}_d)}{\sum_{\substack{ d\prime=0 \\ d\prime \neq d }}^{D} P^{\mathrm rx}_{u,d\prime,k}(\rho^{\mathrm{U}}_u, \rho^{\mathrm{D}}_d)+ \sigma^2_k},
	\label{eq_intro:sinr}
\end{eqnarray}
where $\sigma^2_k$ is the noise power in the $k^{th}$ frequency resource.

By applying the Shannon-Hartley theorem,
the data transmission rate of the $u^{th}$ first responder connected to the BS of the $d^{th}$ UAV in the $k^{th}$ frequency resource can be computed as~\cite{Djordjevic2022}:
\begin{eqnarray}
 	R_{u,d,k}(\rho^{\mathrm{U}}_u, \bm{\rho}^\mathrm{D}) = B_k \log_2(1+\gamma_{u,d,k}(\rho^{\mathrm{U}}_u, \bm{\rho}^\mathrm{D})),
    \label{eq_intro:rate}
\end{eqnarray}
where $B_k$ is the bandwidth of the $k^{th}$ frequency resource.
If a scheduler is used to fairly distribute the available resources among the first responders in the cell, 
such as round-robin~\cite{rasmussen2008round},  
the transmission rate can derived as:
\begin{equation}
    R_u(\rho^{\mathrm{U}}_u, \bm{\rho}^\mathrm{D}) =  \frac{B}{U} \log_2(1+\bar \gamma_{u,d,k}(\rho^{\mathrm{U}}_u, \bm{\rho}^\mathrm{D})),
    \label{eq_intro:mean_rate}
\end{equation}
where $\frac{B}{U}$ indicates the portion of the bandwidth that is allocated to each UE, on average, in the communication.  
Note that the average SINR $\bar \gamma_{u,d,k}$ of the UE in its allocated frequency resources is used as its effective SINR.

It should be noted that,
unlike many other approaches, 
our proposed scheme does not rely on GPS data from UEs for optimization, 
due to the uncertainty of its availability and reliability in emergency situations. 
Instead, 
we assume the use of reference signals, 
and angle of arrival (AoA) estimations over them, 
as proxies. 
In more details, 
we assume that the BS of the $d^{th}$ UAV  can estimate 
---using an antenna array and signal processing--- 
the AoA $\alpha_{u,d}$ of the reference signals received from the $u^{th}$ UE.
To comprehensively evaluate our approach, 
we adopt a two-phase analysis. 
First, we assume perfect AoA estimation by the BS of the $d^{th}$ UAV to establish a performance benchmark, 
allowing us to understand the full potential of our algorithm in ideal conditions. 
Then, we introduce Gaussian noise to simulate real-world imperfections in AoA estimation, 
assessing the algorithm’s robustness and effectiveness under more practical conditions.
\section{Problem Statement}
The objective of this work is to determine, in real-time, the UAV positions $\bm{\rho}^\mathrm{D}$ that maximize the total fair transmission rate \( R_{\text{fair}} \) of first responders.
This rate is defined as the sum of the logarithms of the transmission rates of all UEs~\cite{9878252},
that is:
\begin{equation}
    R_{\text{fair}}(\bm{\rho}^\mathrm{U}, \bm{\rho}^\mathrm{D})  = \sum_{u \in \mathcal{U}}  \log_{10} (R_{u}
     (\rho^\mathrm{U}_u, \bm{\rho}^\mathrm{D})).
     \label{eq_intro:fair_rate}
\end{equation}
This ensures a balanced approach where increases in rates for UEs with lower rates are given more importance compared to those with higher rates, 
thus promoting fairness.

With this in mind,
our optimization problem can then be formally formulated as:
\begin{equation}
    \text{max}_{\bm{\rho}^\mathrm{D}} R_{\text{fair}}(\bm{\rho}^\mathrm{U}, \bm{\rho}^\mathrm{D}).
\end{equation}

The problem's high dimensionality, stochasticity, and non-linearity makes this real-time UAV trajectory optimization highly challenging.
The continuous movement of first responders and fluctuations in the radio channel demand adaptive decision-making,
which traditional optimization techniques struggle to handle effectively.
To address these challenges, 
and given the advantages of PPO highlighted in the introduction,
we adopt a PPO algorithm to dynamically adjust UAV positions.
PPO enables our UAVs to continuously optimize $R_{\text{fair}}(\bm{\rho}^\mathrm{U}, \bm{\rho}^\mathrm{D})$,
adapting to UE mobility patterns and radio condition variations in real-time.
The following section provides a detailed RL formulation and our PPO implementation.

\section{DRL-based PPO Algorithm}

In RL, 
an agent interacts with an environment,
learning to make decisions by receiving feedback in the form of rewards.  
This interaction is formalized through four key components:  
\begin{itemize}
    \item \textbf{States (S):} The possible situations the agent can observe.  
    \item \textbf{Actions (A):} The decisions the agent can take.  
    \item \textbf{Rewards (R):} The feedback the agent receives after taking an action.  
    \item \textbf{Policy ($\pi$):} The strategy the agent follows to decide which action to take in each state.  
\end{itemize}

Like its predecessor TRPO, 
PPO is an on-policy, model-free algorithm,
and belongs to the actor-critic family~\cite{schulman2015trust}.  
It extends the REINFORCE algorithm~\cite{williams1992simple} by incorporating a value function estimator,  
which stabilizes training and improves sample efficiency.  

Specifically,
PPO employs an advantage function to determine how much better a particular action is compared to the average action in a given state.  
This advantage function is defined as:  
\begin{equation}
     \hat{A}_t = r_t - V(s_t), 
\end{equation}  
where \( s_t \) and \( r_t \) represent the state and reward at time step \( t \),  
and \( V(s) \) is the value function.  
This advantage-based approach, 
combined with a value function, 
gives PPO its actor-critic nature,  
allowing simultaneous optimization of both the policy \( \pi \) and the value function \( V(s) \).  

Compared to traditional policy gradient methods~\cite{sutton2018reinforcement},
PPO is designed to address three key challenges:  
\begin{itemize}
    \item PPO improves training stability by using a clipped surrogate objective, 
    which limits the magnitude of policy updates and prevents large, destabilizing changes that can occur in traditional policy gradient methods.
    \item Despite being an on-policy method, 
    PPO is relatively sample-efficient, achieving good performance with fewer environment interactions compared to other RL methods.
    \item PPO also operates in both continuous  states-actions spaces, eliminating the need for discretization and providing a realistic and accurate representation.
\end{itemize}

\subsection{PPO Key Details}

PPO uses a clipped objective to ensure efficiency and stability,
while being easier to implement compared to other methods. 
The algorithm optimizes the policy by implementing a ``surrogate" objective function~\cite{schulman2017proximal}. 
The surrogate objective is given by:
\begin{equation}
    \begin{aligned}
    L^{\text{CLIP}}(\theta) &= \\
    & \hat{\mathbb{E}}_t[\min\big(r_t(\theta)\hat{A}_t, \text{clip}(r_t(\theta), 1-\epsilon, 1+\epsilon)\hat{A}_t\big)],
    \end{aligned}
    \label{eq:clipperd_loss}
\end{equation}
where \(\theta\) represents the weight distribution of DNNs used in the algorithm, 
\( \hat{A}_t \) is the advantage function estimator at time-step \( t \), 
\( \epsilon \) is a hyperparameter controlling the extent of the policy update, 
and \(r_t(\theta)\) measures how much the policy has changed between updates.
Specifically, \(r_t(\theta)\) is the ratio of the probability of taking action \(a_t\) given state \(s_t\) under the current policy \(\pi_\theta\) to the probability of taking that same action under the old policy \(\pi_{\theta_{\text{old}}}\)~\cite{schulman2017proximal}:
\begin{equation}
    r_t(\theta) =\frac{\pi_\theta(a_t \mid s_t)}{\pi_{\theta_{\text{old}}}(a_t \mid s_t)}.
\end{equation}

PPO differs from TRPO by using the clipping mechanism instead of KL divergence.
The clipping ensures that the policy's updates do not deviate too far,
effectively saturating the objective value when the policy's update becomes too large. 
This discourages excessive changes to the policy by taking the minimum of the clipped and unclipped objectives.
Additionally, PPO incorporates generalized advantage estimation (GAE)~\cite{schulman2015high} to refine the advantage function. 
GAE applies a discount factor \( \lambda \) to balance the consideration of immediate and future rewards, 
which results in more efficient policy learning. 
This enables PPO to navigate complex environments, 
such as those encountered in UAV flights for first responders.


\subsection{Proposed PPO Implementation}

Without loss of generality,
considering a single UAV through subsequent experiment, 
we aim to analyze how the PPO optimizes the $d^{th}$ UAV position to serve the UEs of the first responders efficiently.
To this end, 
we define the state-space environment of our problem as follows:
\begin{equation}
    \tilde{\rho} = [\rho_{d,t}^{D}, \rho_{d,t-1}^{D}, \dots, \rho_{d,t-M}^{D}],
\end{equation}
\begin{equation}
    \tilde{\gamma}_u = [\tilde{\gamma}_{u,t}, \tilde{\gamma}_{u,t-1}, \dots, \tilde{\gamma}_{u,t-M}] ,
\end{equation}
\begin{equation}
    \mu_{\alpha} = [\mu_{\alpha,t}, \mu_{\alpha,t-1}, \dots, \mu_{\alpha,t-M}],
\end{equation}
\begin{equation}
    \sigma_{\alpha} = [\sigma_{\alpha,t}, \sigma_{\alpha,t-1}, \dots, \sigma_{\alpha,t-M}],
\end{equation}
\begin{equation}
    \text{S} = [\tilde{\rho}, \tilde{\gamma}_u, \mu_{\alpha}, \sigma_{\alpha}],
\end{equation}
where $\tilde{\rho}$ is the time-dependent vector of the positions of $d^{th}$ UAV,
$\tilde{\gamma}_u$ is the time-dependent vector of SINR measurements of the $u^{th}$ UE connected to the BS of the $d^{th}$ UAV, 
and \(\mu_{\alpha}\) and \(\sigma_{\alpha}\) denote the mean and standard deviation of the AoAs \(\alpha_{u,d}\ \forall u\) of the reference signals transmitted by the $u^{th}$ UEs of the first responders and received by the BS of the $d^{th}$ UAV. 

Putting all these variables together, 
S is the state vector formed by Pos, $\tilde{\gamma}_u$, $\mu_{\alpha}$, and $\sigma_{\alpha}$. 
Note that the parameter $M$ represents the memory length, 
i.e. we store the last $M$ values in each vector. 

As we have a continuous action space, 
the agent has two variables to choose from: 
direction and movement magnitude. 
The direction is defined by the angle $\alpha_d$, 
and the magnitude is defined by the distance $r$. 
Hence, the action space is defined as $\mathcal{A} = \{(\alpha_d, r)\}$, 
where $\alpha_d \in [-180, 180)$ and $r \in [0, r_{max}]$. 
Here, $r_{max}$ is the maximum distance the UAV can travel in one action, 
and the angle $\alpha_d$ is defined with respect to the east. 
It is important to note that our implementation leverages the PPO's ability to work with continuous action spaces. 
By avoiding discretization, 
the agent can explore the full continuum of possible positions, 
enabling it to find optimal locations that maximize system throughput with high precision. 
This is particularly beneficial in our scenario, 
where small adjustments in UAV position can significantly impact network performance.

Through our proposed framework, 
the reward function is designed to maximize the total fair transmission rate $R_{\text{fair}}$, 
as introduced earlier. 
To ensure consistency and comparability of the reward values during training, 
a min-max normalization is applied. 
This normalization centers the reward values and scales them, 
improving the stability and efficiency of the RL algorithm.

\section{Numerical Results and Discussion}

This section evaluates our UAV-based BS flights algorithm across various dynamic emergency scenarios, focusing on convergence, learned flights strategies, and network performance, particularly the total fair transmission rate $R_{fair}$.

\subsection{Experimental Setup}

\subsubsection*{Scenario}

To evaluate our algorithm's effectiveness, 
we consider a 200m × 200m area,
and that the BS of $d^{th}$ UAV operates a bandwidth $B=$10\,MHz at frequency $f=$ 2\,GHz band.
We use the system model presented in Section \ref{sec:system_model},
embracing the 3GPP UMa model in TR25.814.

\subsubsection*{UE Mobility Models}
We implement various UE mobility patterns within the designated area, increasing in complexity, where UEs reflect off boundaries upon contact. The UE mobility patterns are as follows:
\begin{itemize}
    \item \textit{Static UEs (No Move)}: A cluster of 10 UEs in a 10m × 10m square at the center, representing stationary scenarios.
    
    \item \textit{Linear Motion (Straight Walk)}: The same cluster moving in a straight line at 8 m/s, simulating unified group movement 
    ---the direction of the movement is randomly selected.
    
    \item \textit{Circular Motion}: The same cluster initialized at $x \in [0, 10]$, $y \in [-190, -200]$, 
    moving in a circular path (radius 200m, center at origin), 
    representing a more complex search pattern.
    
    \item \textit{Crossed Linear Motion}: Two clusters of 5 UEs each, 
    moving perpendicularly (90\textdegree\ gap) or in opposite directions (180\textdegree\ gap), simulating multiple independent groups.

    \item \textit{Random Hotspot and Mixed Movement}: Two groups of 5 UEs each: one cluster moving uniformly and another set of independently moving UEs. 
    Velocities are randomly assigned within 0 to 8 m/s, 
    simulating both coordinated and unpredictable movements in dynamic scenarios.
\end{itemize}
\subsubsection*{PPO Algorithm Configuration}

Our proposed algorithm based on PPO is trained using the following key parameters and configuration. 
The PPO algorithm is trained for 12,000 episodes, 
with each episode consisting of 128 frames. 
We employed a learning rate of 3e-4, 
and implemented three hidden layers in both the actor and critic networks, 
each containing 128 neurons. 
The algorithm used a discount factor $\gamma$ of 0.99 and a GAE parameter $\lambda$ of 0.95, 
balancing immediate and future rewards effectively. 
To ensure stable learning, 
we applied gradient clipping with a maximum norm of 1.0, 
and used a clip $\epsilon$ of 0.2 for the surrogate objective. 
Our model incorporated a memory size $M$ of 4, 
enabling the agent to have an idea of what the UE movement is without having the UE position directly.

\subsubsection*{Results and Discussion}

\begin{figure}[h]
\centering
\includegraphics[width=0.49\textwidth]{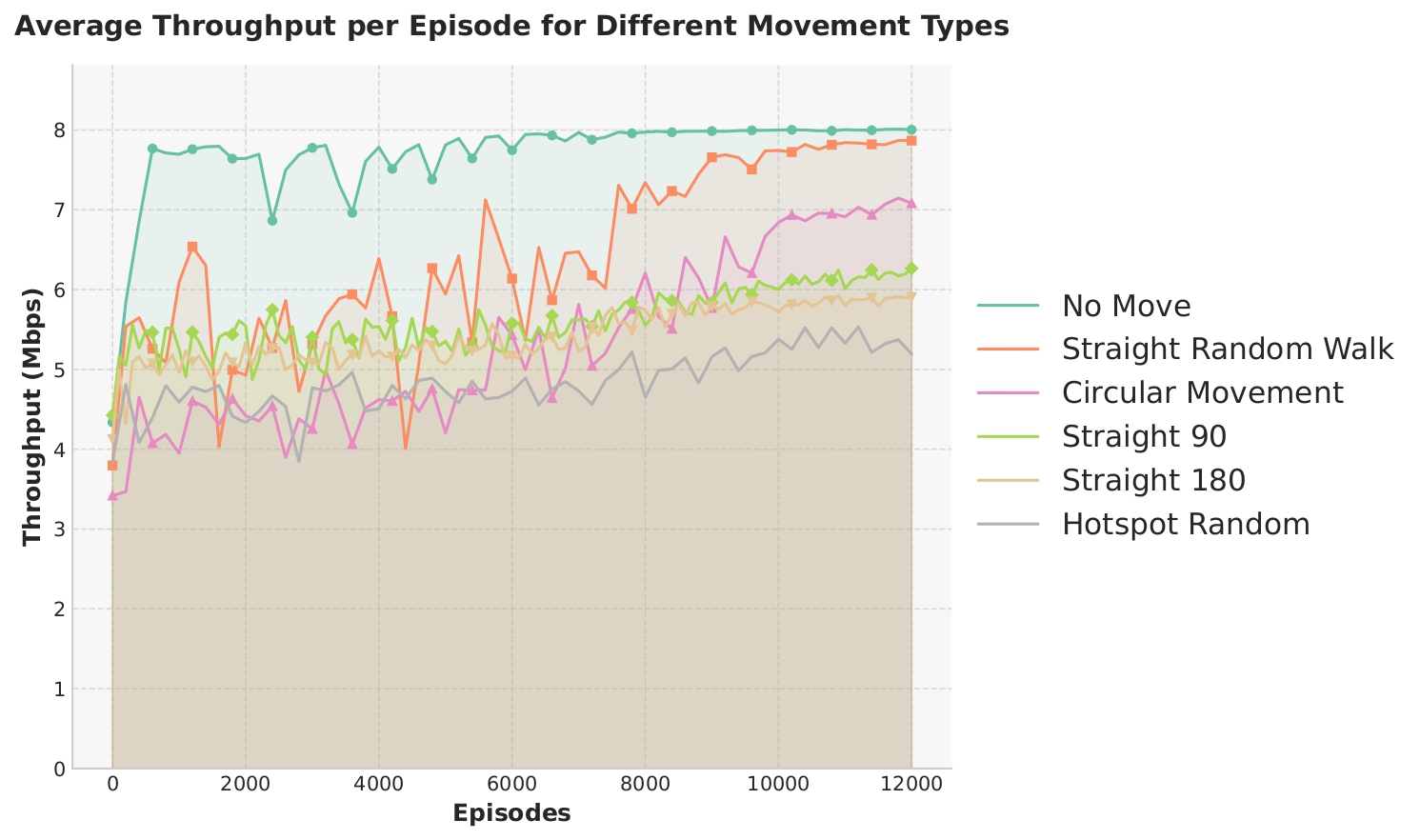}
\caption{Average throughput per evaluation episode for different movement types}
\label{throughput_analysis}
\end{figure}

Figure~\ref{throughput_analysis} illustrates the average throughput per episode during the evaluation phase for the various movement scenarios.
For reference, 
given the configuration,
the maximum achievable UE rate is around 8\,Mbps.
The results reveal distinct trends for each movement type, highlighting the algorithm's adaptability and effectiveness across diverse scenarios. In scenarios with a single cluster of first responders (``No Move'', ``Straight Random Walk'', and ``Circular Movement''), the agent consistently achieves throughputs exceeding 7\,Mbps. This performance is maintained even with randomly varying speeds, demonstrating the agent's robust ability to optimize its position based solely on reference signal information. It is worth noting that the complex scenarios resulted in performance degradation and failed to achieve the maximum rate.
For scenarios involving two clusters of first responders (``Straight 90'' and ``Straight 180''), the agent effectively learns to optimize its position, progressively improving throughput over the episodes. This improvement indicates the algorithm's capacity to handle more complex spatial distributions of UEs, balancing performance between multiple groups. The ``Straight 180" scenario is the most challenging here, as the UE clusters get farther apart compared to the ``Straight 90" scenario, resulting in lower performance.
The ``Hotspot Random'' scenario, representing the most complex movement pattern, shows a gradual improvement in throughput, albeit with more variability. This scenario challenges the agent with both clustered and dispersed UEs moving at random velocities, yet the algorithm still manages to enhance performance over time.
These results collectively underscore the algorithm's adaptability across diverse movement patterns and its capacity to enhance network performance using only information derived from reference signals without relying on UE location information.

To further evaluate the effectiveness of our proposed approach, 
we compare the performance of our PPO-based agent against another baseline scenario where the UAV is statically positioned at the center of the environment. 
Table~\ref{tab:comparison} presents this comparison for each movement scenario.
\begin{table}[htbp]
\centering
\caption{Comparison of Average Throughput: PPO vs. Static}
\label{tab:comparison}
\small
\begin{tabular}{@{}lcccc@{}}
\toprule
\textbf{Scenario} & \textbf{PPO} & \textbf{Static} & \textbf{Diff.}&\textbf{Gain} \\
& \textbf{(Mbps)} & \textbf{(Mbps)} & \textbf{(Mbps)}& \textbf{(\%)} \\
\midrule
No Move & 8.00 & 7.95 &+0.05 &2\% \\
Straight Random & 7.87 & 4.59 & +3.28&71\% \\
Circular & 7.08 & 3.59 & +3.49&97\% \\
Straight 90\textdegree & 6.27 & 5.33 & +0.94&17\% \\
Straight 180\textdegree & 5.91 & 5.31 & +0.60&11\% \\
Hotspot Random & 5.19 & 4.77 & +0.42&8\%\\
\bottomrule
\end{tabular}
\end{table}
The results for the ``Straight 90°" and ``Straight 180°" scenarios are particularly noteworthy, with the PPO algorithm outperforming the static solution by 17.64\% and 11.30\%, respectively.
This improvement may be surprising, as one might expect a centrally located static UAV to be an optimal solution for two groups moving in perpendicular or opposite directions. However, the fact that our PPO-based approach, which takes radio propagation conditions and network performance into account through the reward function, identifies a strategy that surpasses this geometrically optimal static position suggests that the algorithm is effectively leveraging its mobility to optimize coverage in a dynamic and non-obvious way.

Furthermore, we analyze the convergence when Gaussian noise is added to the AoA estimation. 
Table~\ref{tab:throughput_std_straight_random} presents the results of this simulation for the ``Straight Random" movement scenario.
\begin{table}[htbp]
    \centering
    \caption{Throughput values for the ``Straight Random" scenario with different noise levels}
    \label{tab:throughput_std_straight_random}
    \small
    \begin{tabular}{@{}lcccccccc@{}}
        \toprule
        \textbf{STD} & 0 & 1 & 5 & 100 & 50 & 100 \\
        \midrule
        \textbf{PPO (Mbps)} & 7.87 & 7.84 & 7.82 & 7.80 & 7.70 & 7.16 \\
        \bottomrule
    \end{tabular}
\end{table}
The results demonstrate the algorithm's robustness to noise in AoA estimation. Even with a significant noise level (STD = 100), the throughput remains high at 7.16 Mbps, only a 0.71 Mbps decrease from the noise-free scenario. This resilience to noise underscores the practical viability of our approach in real-world deployments, where perfect AoA estimations may be unlikely. The algorithm's ability to maintain high throughput even with noisy AoA measurements highlights its potential for reliable performance in challenging emergency communication scenarios.

\balance

\section{Conclusion}

This paper studies the UAV-based BS problem in emergency communications by emphasizing the importance of optimal UAV flights to enhance network performance. It introduces a continuous action space DRL approach using PPO, allowing UAVs to dynamically select both direction and movement magnitude. The key innovation is the use of widely available reference signals to drive the state-action space, improving model realism in scenarios where GPS or accurate UE location data is unavailable. The results confirm the algorithm's adaptability and effectiveness in maintaining comprehensive coverage across various scenarios, marking a significant advancement in real-time UAV flights for improved emergency response.
As a future work, 
we will consider deploying multiple UAVs as aerial BSs to further enhance network coverage and reliability,
and explore advanced coordination strategies. 


\end{document}